\documentclass[12pt]{l4dc2023}

\usepackage{wrapfig}
\usepackage{times}
\usepackage{lipsum}
\usepackage{amsmath} % allows further math commands
\usepackage{amssymb} % allows further math symbols
\usepackage{amsfonts} % allows further math symbols
\usepackage{mathtools} % allows further math commands
\usepackage{algorithm}		
\usepackage{enumitem}
\usepackage[noend]{algcompatible}

\usepackage{tikz}
\usepackage{pgfplots}
\usepackage{stackengine}
\usepackage{soul}
\usepackage{multirow}
\usepackage[normalem]{ulem}
\tikzstyle{block} = [draw, rectangle, 
    minimum height=3em, minimum width=5em]
\tikzstyle{input} = [coordinate]
\tikzstyle{output} = [coordinate]
\tikzstyle{pinstyle} = [pin edge={to-,thin,black}]
\usetikzlibrary{shapes.geometric,shapes.arrows,decorations.pathmorphing,calc}
\usetikzlibrary{matrix,chains,scopes,positioning,arrows,fit}
\usepackage{soul}
\usepackage{cases}

\usepackage{setspace}

\usepackage{bm}           % to make greek letters bold
\usepackage{booktabs}

\setlist{noitemsep} 		          % remove spacing between bullet/numbered list elements

\newcommand{\T}{^{\mathsf{T}}}

\newcommand{\B}[1]{\if#1\relax\bm{#1}\else\mathbf{#1}\fi} % bold text THIS IS BUGGED IN TABLES DOES NOT WORK
\newcommand{\R}[1]{\mathrm{#1}}						      % regul. text
\newcommand{\C}[1]{\mathcal{#1}}
\newcommand{\BB}[1]{\mathbb{#1}}
\newcommand{\norm}[1]{\left\lVert #1 \right\rVert}

\usepackage[prependcaption,colorinlistoftodos]{todonotes}

\title[Control-Tutored DQN]{CT-DQN: Control-Tutored Deep Reinforcement Learning}
\usepackage{times}
\author{%
 \Name{Francesco De Lellis} \Email{francesco.delellis@unina.it}\\
 \addr University of Naples Federico II, Italy%
 \AND
 \Name{Marco Coraggio} \Email{marco.coraggio@unina.it}\\
 \addr Scuola Superiore Meridionale, Italy%
 \AND
 \Name{Giovanni Russo$^*$} \Email{giovarusso@unisa.it}\\
 \addr University of Salerno, Italy%
 \AND
 \Name{Mirco Musolesi$^*$} \Email{m.musolesi@ucl.ac.uk}\\
 \addr University College London, UK, and University of Bologna, Italy%
 \AND
 \Name{Mario di Bernardo$^*$} \Email{mario.dibernardo@unina.it}\\
 \addr University of Naples Federico II, Italy, and Scuola Superiore Meridionale, Italy%
}
\begin{document}
\maketitle

\begin{abstract}%
One of the major challenges in Deep Reinforcement Learning for control is the need for extensive training to learn the policy.
Motivated by this, we present the design of the Control-Tutored Deep Q-Networks (CT-DQN) algorithm, a Deep Reinforcement Learning algorithm that leverages a control tutor, i.e., an exogenous control law, to reduce learning time.
The tutor can be designed using an approximate model of the system, without any assumption about the knowledge of the system's dynamics. 
There is no expectation that it will be able to achieve the control objective if used stand-alone. 
During learning, the tutor occasionally suggests an action, thus partially guiding exploration.
We validate our approach on three scenarios from OpenAI Gym: the inverted pendulum, lunar lander, and car racing.
We demonstrate that CT-DQN is able to achieve better or equivalent data efficiency with respect to the classic function approximation solutions.
\end{abstract}


%----------------------------------
\section{Introduction}

% MOTIVATION
The design of controllers based on training from data via Reinforcement Learning (RL) is a fascinating area, which is also increasingly gaining popularity.
This paradigm is particularly suitable for scenarios in which we do not have any prior knowledge of the system dynamics \citep{bertsekas1996neuro,Sutton1998,nian2020review}. 
At the same time, in order to deal with large state spaces, neural approximators are now widely adopted. These solutions are usually referred to as {\em Deep Reinforcement Learning (DRL)} \citep{hornik1989multilayer,mnih2015human,lillicrap2015continuous}.
Control algorithms based on DRL have shown impressive performance in different application fields, including the  control of plasma in nuclear fusion \citep{degrave2022magnetic} and that of microbial cultures in bioreactors \citep{treloar2020deep}. However, a crucial challenge for these algorithms is that they typically require extensive training.
To tackle these challenges, we propose a control framework combining DRL algorithms and feedback controllers. 

Indeed, in recent years, classical control theoretical tools and RL have been intertwined in a number of ways. For example, in \cite{rathi2020driving}, Model Predictive Control (MPC) was used in state-space regions where a model of the dynamics is available, while tabular Q-learning was used in the other regions.
Instead in \cite{zanon2020safe}, a RL algorithm is used to vary the parameters of the model and the objective function used by a MPC.
The authors of \cite{abbeel2006using} propose a policy gradient algorithm that performs updates on the policy using data generated by an approximate Markov decision process model in combination and through exploration of the environment.
In \cite{gu2016continuous}, a variant of the Q-learning algorithm (normalized advantage functions) is discussed; the authors shows that their solution is able to accelerate the learning process by using local linear models fitted iteratively with exploration data.
More in general, model-based RL techniques have been developed to learn the system dynamics; the model is then used to perform simulated roll-outs, which generate new data for learning \citep{sutton1991dyna}. An example is \cite{deisenroth2011pilco}, in which model fitting is performed using  Gaussian processes.
However, these model-based techniques may introduce biases in the learning process as part of the data is not generated by the actual system.
Conversely, in \cite{de2020tutoring, pmlr-v168-lellis22a} the authors propose Control-Tutored Reinforcement Learning, which relies on the introduction of a \emph{beneficial} bias in the exploration process to speed up learning.
This bias has the form of ``suggestions'' from a control law---which we call \emph{tutor}---based on approximate modeling of the system dynamics.
Moreover, in \cite{pmlr-v168-lellis22a}, with the objective of stabilizing an inverted pendulum, the authors integrate Q-learning with a simple tutor designed to capture a very limited description of the dynamics. This approach leads to a significant reduction in the learning time, without compromising the effectiveness of the learnt policy. 
However, the algorithm employs tabular RL, which is suitable only for problems with limited state and action spaces.

% CONTRIBUTION
The key contributions of this paper can be summarized as follows. We introduce Control-Tutored Deep Q-Networks (CT-DQN), an algorithm combining DQN \citep{mnih2015human} and control tutors, to show how the latter can be effectively applied in DRL. We discuss its design show its effectiveness via numerical validations. Namely, we test our algorithm in three representative OpenAI Gym scenarios of increasing complexity to stress the performance of our approach.
Then, by comparing our results to a classical DQN, we show that, even if the tutor is implemented through a very simple control law and designed using a {\em rough} approximation of the underlying dynamics, learning time can be reduced significantly, thus improving data efficiency of the learning process. 

%------------------------------------
\section{Control-Tutored Deep Q-Networks}
\label{sec:CT-DQN}
In the following, we denote random variables by capital letters and their realization with lower case letters; $\BB{E}$ is the expectation operator.

%---------------------------------------
\subsection{Problem Formulation}
\label{sec:problem_formulation}

Following \cite{pmlr-v168-lellis22a, de2020tutoring}, we consider a discrete time dynamical system affected by noise, of the form
\begin{equation}\label{eq:dynamical_system}
    X_{k+1} = f(X_k, U_k, W_k),
    \quad x_0 = \tilde{x}_0,
\end{equation}
where $k \in \BB{N}_{\ge 0}$ is discrete time,
$X_k \in \C{X}$ is the state at time $k$, with
$\C{X}$ being the state space, 
$\tilde{x}_0 \in \C{X}$ is the initial condition, 
$U_k \in \C{U}$ is the {control input} (or {action}), and $\C{U}$ is the set of feasible inputs; $W_k$ is a random variable representing noise, with values in a set $\C{W}$, and $f : \C{X} \times \C{U} \times \C{W} \rightarrow \C{X}$ is the system's {dynamics}.%

We consider the problem of learning a policy $\pi : \C{X} \rightarrow \C{U}$ 
to solve the following {sequential decision making problem, see e.g., \citep{GARRABE202281},} with finite time horizon $N \in \BB{N}_{>0}$: 
\begin{subequations}\label{eq:rl_problem_statement}
\begin{align}
    \max_{\pi}& \ \ \BB{E}[J^{\pi}],\\ 
    \text{s.t.}
    & \ \ X_{k+1} = f(X_k,U_k,W_k),
        \quad k \in \{ 0, \dots, N-1 \},\\
    &\ \ U_k = \pi(X_{k}),  
        \quad k \in \{ 0, \dots, N-1 \},\\
    &\ \ x_0 \ \text{given},
\end{align}
\end{subequations}
where
$ J^{\pi} = r_N(X_{N}) + \sum_{k=1}^{N} r(X_{k}, X_{k-1}, U_{k-1})$
is the \emph{cumulative reward}, with $r : \C{X} \times \C{X} \times \C{U} \rightarrow \BB{R}$ being the \emph{reward} received by the learning agent when entering the next state after taking the selected action and $r_N:\C{X} \rightarrow \BB{R}$ being the \emph{final reward}.

%---------------------------------------
\subsection{Policy Design}
\label{sec:policy_design}

During the learning phase, the control input $U_k$ is chosen either as the value proposed by some RL policy ($\pi^{\R{rl}} : \C{X} \rightarrow \C{U}$), with probability $\beta \in (0, 1)$, or as the one proposed by a control law (i.e., \emph{tutor}; $\pi^{\R{c}} : \C{X} \rightarrow \C{U}$).
Hence, $\pi$ in \eqref{eq:rl_problem_statement} is given by
\begin{equation}\label{eq:ctrl}
    \pi(x) =
    \begin{dcases}
        \pi^{\R{rl}}(x), & \text{with probability } \beta,\\
        \pi^{\R{c}}(x), & \text{with probability  } 1 - \beta.
    \end{dcases}
\end{equation}

Next, we explain how we selected $\pi^{\R{rl}}$ and $\pi^{\R{c}}$ in \eqref{eq:ctrl}.
Specifically, $\pi^{\R{rl}}$ is learnt through an $\epsilon$-greedy DQN policy \citep{mnih2015human}.
Thus, we have
\begin{subnumcases}{\pi^{\R{rl}}(x) =
\label{eq:rl_policy}}
        \arg \max_{u\in \C{U}}Q(x, u),
            & $\text{with probability } 1 - \epsilon^{\R{rl}}$,
            \label{eq:rl_policy_q}\\
        u \sim \R{rand}(\C{U}),
            & $\text{with probability } \epsilon^{\R{rl}}$,
            \label{eq:rl_policy_rand}
\end{subnumcases}
\noindent with $\epsilon^{\R{rl}} \in (0, 1)$, and $Q : \C{X} \times \C{U} \rightarrow \BB{R}$ being the \emph{state-action value function} \citep{Sutton1998}.
DQN uses Deep Neural Networks to iteratively  approximate the function $Q$; it is among the most popular implementations of DRL and can be used also for continuous state spaces $\C{X}$.
Differently from tabular \emph{Q-learning} \citep{Watkins1992}, there are currently no guarantees of convergence towards the optimal policy for DQN \citep{pmlr-v120-yang20a}, although its effectiveness is supported by strong empirical evidence \citep{mnih2015human, mnih2013playing}.

To select the \emph{control tutor} policy $\pi^{\R{c}}$ in \eqref{eq:ctrl}, we assume to have a feedback controller $g : \C{X} \rightarrow \C{U}$, designed with limited information%
\footnote{See Section \ref{sec:evaluation} for practical examples.} about the dynamical system described by~\eqref{eq:dynamical_system}. 
Then, letting $\epsilon^{\R{c}} \in (0, 1)$, we select 
\begin{subnumcases}
    {\pi^\R{c}(x) =
    \label{eq:tutor_policy}}
        g(x),
            & $\text{with probability } 1 - \epsilon^{\R{c}}$,
            \label{eq:tutor_policy_g}\\
        u \sim \R{rand}(\C{U}),
            & $\text{with probability } \epsilon^{\R{c}}$.
            \label{eq:tutor_policy_rand}
\end{subnumcases}

In conclusion, combining \eqref{eq:ctrl}, \eqref{eq:rl_policy} and \eqref{eq:tutor_policy}, we have that
action \eqref{eq:rl_policy_q} is taken with probability $\beta (1 - \epsilon^{\R{rl}})$, action \eqref{eq:tutor_policy_g} is taken with probability $\omega \coloneqq (1 - \beta) (1 - \epsilon^{\R{c}})$ and the random action with probability $\beta \epsilon^{\R{rl}} + (1-\beta) \epsilon^{\R{c}}$.

%--------------------------------------------------
\section{Metrics}
\label{sec:metrics}

In all scenarios we consider, each study is repeated in $S$ independent \emph{sessions}, each composed of $E$ \emph{episodes}, which are simulations lasting $N$ time steps.
The weights of the neural networks in DQN are carried over from one episode to the next, and re-initialized at each session.
An episode can end earlier if a (scenario-specific) \emph{terminal condition} is met, and we denote by $J_e^{\pi}$ the cumulative reward (see § \ref{sec:problem_formulation}) in episode $e$.
As usual, maximizing $J^{\pi}$ in § \ref{sec:problem_formulation} amounts to fulfilling some problem-specific \emph{goal}: we define the \emph{goal condition} $c_{\R{g}}$ as a Boolean variable that is true if and only if the goal is achieved in an episode.
Next, we define three metrics to assess learning performance.

\begin{definition}[Learning metrics]\label{def:learning_metrics}
    (i) The \emph{average cumulative reward} is
    $J_{\R{avg}}^{\pi} \coloneqq \frac{1}{E} \sum_{e = 1}^E J_e^{\pi}$.
    (ii) The \emph{terminal episode} $E_{\R{t}}$ is the smallest episode such that $c_{\R{g}}$ is true for all $e \in \{E_{\R{t}}- 10, \dots, E_{\R{t}}\}$.
    (iii) The \emph{average cumulative reward after terminal episode} is
    $J_{\R{avg}, \R{t}}^{\pi} \coloneqq \frac{1}{E_{\R{t}}} \sum_{e = E_{\R{t}}}^E J_e^{\pi}$.
\end{definition}

$J_{\R{avg}}^{\pi}$ is often used in RL \citep{duan2016benchmarking, wang2019benchmarking}; $E_{\R{t}}$ is used to assess the effective duration of the learning phase, and consequently data efficiency; $J_{\R{avg}, \R{t}}^{\pi}$  quantifies the quality of the controller, once the learning phase is completed.
Next, we define three metrics inspired by those commonly used in control theory, to assess the transient and steady-state performance.
Let $\pi_{\R{greedy}}(x) = \arg \max_{u \in \C{U}} Q(x, u)$ be the \emph{greedy policy}.
Moreover, when the goal is (or entails) achieving some \emph{goal state} $x^* \in \C{X}$ (or region containing $x^*$), we say the goal is a \emph{regulation problem}.

\begin{definition}[Control metrics]\label{def:control_metrics}
% \begin{itemize}
%     \item
    (i) The \emph{cumulative reward} (see § \ref{sec:problem_formulation}) \emph{obtained following} $\pi_{\R{greedy}}$ is $J^{\pi_{\R{greedy}}}$; 
    % \item 
    (ii) in an episode, the \emph{settling time} $k_{\R{s}}$ is a time instant such that the goal is achieved or a related task is completed (defined uniquely in each scenario, when possible);
    % \item
    (iii) in regulation problems, the \emph{steady state error} is $e_{\R{s}} \coloneqq \frac{1}{N - k_{\R{s}} + 1} \sum_{k = k_{\R{s}}}^{N}
        \norm{x - x^*}$.        
% \end{itemize}
\end{definition}

%----------------------------------------
\section{Evaluation}
\label{sec:evaluation}

%In the following, 
We assess the performance of the CT-DQN algorithm \eqref{eq:ctrl}-\eqref{eq:rl_policy}-\eqref{eq:tutor_policy} on three representative case studies from the OpenAI gym suite \citep{GYM, brockman2016openai}, i.e., the inverted pendulum \cite{GYM_pendulum}, lunar lander \cite{GYM_lander} and car racing \cite{GYM_car}.
The inverted pendulum was selected as it is a classical nonlinear benchmark problem in control theory.
Lunar lander was chosen as it represents a harder control problem with multiple input and outputs (MIMO) and in which certain regions of the state space must be avoided.
Car racing was selected as it is a tracking problem where the state is observable as a matrix of pixels, rather than measured physical quantities.

\subsection{Inverted Pendulum} 
\label{sec:inverted_pendulum}
%-------------------------------------
\paragraph{Environment description and control goal.}
A rigid pendulum, subject to gravity, must be stabilized to its upward position. 
The states are the pendulum's angular position and velocity; $\C{X} = [-\pi, \pi] \times [-8, 8]$; $[0 \ \ 0]\T$ and $[\pi \ \ 0]\T$ correspond to the upward unstable position and the downward stable position, respectively; 
%$x_{\R{max}} \coloneqq \norm{[\pi \ \ 8]}$ is half the \emph{diameter} of $\C{X}$;
the initial position is always $x_0 = [\pi, 0]\T$.
The control input is a torque at the joint, with $\C{U}$ being discrete.
Further details are reported in \cite{GYM_pendulum} and omitted here for brevity.
We set $S=3$, $E=100$, and $N=400$ (see § \ref{sec:metrics}).
The control goal is a regulation problem, with $x^* = [0 \ \ 0]\T$.
The goal condition $c_{\R{g}}$ is true in an episode if
$
    \exists \bar{k} \in [0, N-100] :
    \norm{x_k - x^*} \le 0.05 \norm{[\pi \ \ 8]}, 
    \forall k \in [ \bar{k}, N]
$;
the settling time $k_{\R{s}}$ is the smallest of such $\bar{k}$.

%----------------------------------
\paragraph{Control tutor design.}
Assume we know a linearized dynamics of the pendulum, approximating $f$ in \eqref{eq:dynamical_system} close to the upward equilibrium position $x^*$, namely 
$\hat{f}\left(x_{k}, v_{k}\right) = A x_{k} + B v_{k}$,
where 
$A = \left[ \begin{smallmatrix} 
0  &  1 + T  \\
3 T g /2 l  &  1
\end{smallmatrix} \right]$
and 
$B = \left[ \begin{smallmatrix} 
0   \\
T/I
\end{smallmatrix} \right]$,
{with $T=0.05$ s being the sampling time, $l = 1$ m being the rod length and $I = ml^2/3$ being the moment of inertia of the rod}.
From this model, using a pole placement technique, we synthesize the linear feedback controller $v_k = - [5.83 \ \ 1.83] x_k$, which can stabilize $x^*$ only locally.
Then $g(x)$ in \eqref{eq:tutor_policy} is obtained by projecting $v_k \in \BB{R}$ in $\C{U}$ (which is discrete).
Note that this controller, if used on its own, is unable to swing up the pendulum from its downward asymptotically stable position.

%----------------------------------
\paragraph{Numerical results.}

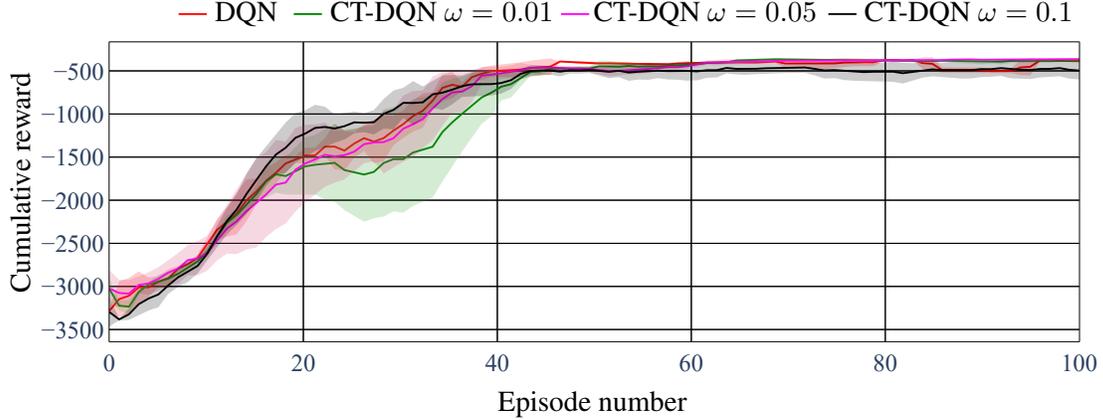
\begin{figure}[t]
\centering
\begin{minipage}{\textwidth}
\begin{tikzpicture}
  \node (img)  {\includegraphics[width=0.98\textwidth]{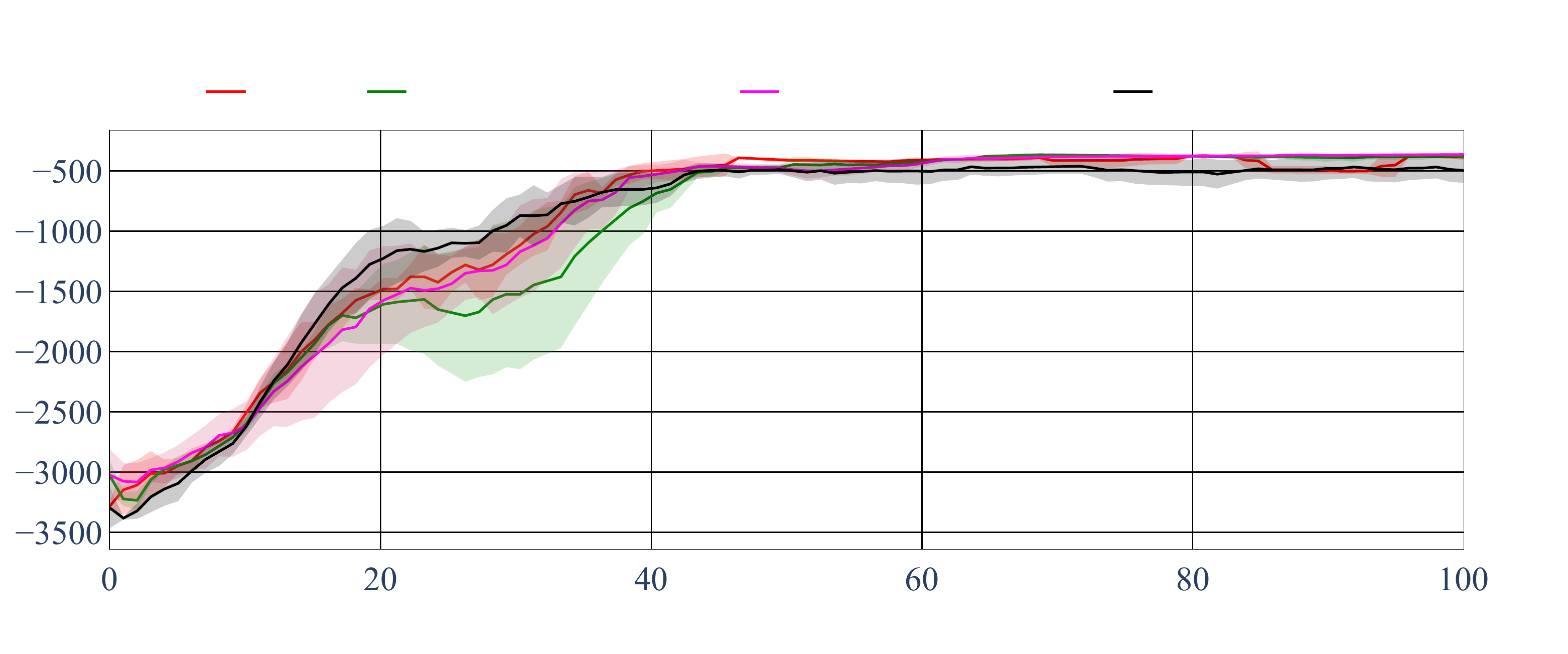}};
  \node[above=of img, node distance=0cm,xshift=-4.6cm,yshift=-2.3cm,font=\color{black}] {DQN};
  \node[above=of img, node distance=0cm,xshift=-2cm,yshift=-2.3cm,font=\color{black}] {CT-DQN $\omega = 0.01$};
  \node[above=of img, node distance=0cm,xshift=1.5cm,yshift=-2.3cm,font=\color{black}] {CT-DQN $\omega =0.05$};
  \node[above=of img, node distance=0cm,xshift=5cm,yshift=-2.3cm,font=\color{black}] {CT-DQN $\omega =0.1$};
  \node[below=of img, node distance=0cm, yshift=1.6cm,font=\color{black}] {Episode number};
  \node[left=of img, node distance=0cm, rotate=90, anchor=center,yshift=-1cm,font=\color{black}] {Cumulative reward};
 \end{tikzpicture}
\end{minipage}
\caption{Cumulative reward per episode $J^\pi_\R{e}$ for the inverted pendulum problem.
The reward curves were averaged with a moving window of 10 samples taken on the left.
Then mean (solid curves) and standard deviations (shaded areas) are taken across sessions.
}
\label{fig:pendulum_rewards}
\end{figure}

Fig.~\ref{fig:pendulum_rewards} shows that CT-DQN (with different values of the switching probability $\omega$) and DQN have comparable performance during the learning phase.
Indeed, in Tab.~\ref{tab:learning_metrics}, a Welch's t-test reveals no statistically significant difference between the two. 
In Tab.~\ref{tab:control_metrics}, we report the control metrics assessed after a training of $50$ episodes (larger than $E_{\R{t}}$ for all cases, meaning learning is considered complete), and observe similar control performance, without statistically significant differences.
Hence, in this scenario, under all metrics considered, CT-DQN and DQN have comparable performance.
We believe this happens because the state and action spaces are small, and DQN is already able to learn quickly, not needing additional aid from the tutor.
%Differently, 
In Sections \ref{sec:lunar_lander} and \ref{sec:car_racing}, we show how the tutor can improve learning performance when the state and action spaces are larger.

\begin{table}[t]
\centering
\begin{tabular}{l|lll}
    \toprule
    Algorithm & $E_{\R{t}}$ & $J_{\R{avg}}^{\pi}$ & $J_{\R{avg,t}}^{\pi}$ \\
    \midrule
    \multicolumn{4}{c}{\textit{Inverted pendulum}}\\
    \noalign{\smallskip}
    DQN & $38\pm3$ & $-837.8\pm90.6$ & $-422.8\pm16.9$\\
    CT-DQN ($\omega = 0.01$) & $40\pm4$ & $-906\pm110.4$ & $-397.3\pm8.9$\\
    CT-DQN ($\omega = 0.05$) & $39\pm4$ & $-856.5\pm111.9$ & $-403.1\pm12.5$ \\
    CT-DQN ($\omega = 0.1$) & $38\pm6$ & $-858.9\pm43.3$ & $-507.1\pm68.6$ \\
    \midrule
    \multicolumn{4}{c}{\textit{Lunar lander}}\\
    \noalign{\smallskip}
    DQN & $665\pm28$ & $60.1\pm1.9$ & $231.6\pm5.2$\\
    CT-DQN ($\omega=0.01$) & $\mathbf{456\pm29}$ & $\mathbf{135.5\pm25.4}$ & $232.5\pm0.9$\\
    CT-DQN ($\omega=0.05$) & $\mathbf{324\pm90}$ & $\mathbf{155.7\pm25.9}$ & $230\pm9.4$ \\
    CT-DQN ($\omega=0.1$) & N.A. & $\mathbf{100\pm13.6}$ & N.A.\\
    \midrule
    \multicolumn{4}{c}{\textit{Car racing}}\\
    \noalign{\smallskip}
    DQN & - & $363.7 \pm 13.7$ & - \\
    CT-DQN ($\omega = 0.05$) & - & $\mathbf{451.7 \pm 0.4}$& -\\
    \bottomrule
\end{tabular}
\caption{Learning metrics (Def. \ref{def:learning_metrics}) for the scenarios in § \ref{sec:evaluation}.
Means and standard deviations across sessions are reported, when $S > 1$.
Values that are statistically significantly different from those of DQN are in bold (according to Welch’s t-test with $p$-value less than $0.05$).
}
\label{tab:learning_metrics}
\end{table}

\begin{table}[t]
\centering
\begin{tabular}{l|lll}
    \toprule
    Algorithm & $k_{\R{s}}$ & $e_{\R{s}}$ & $J^{\pi_{\R{greedy}}}$ \\
    \midrule
    \multicolumn{4}{c}{\textit{Inverted pendulum}}\\
    \noalign{\smallskip}
    DQN & $66\pm3$ & $0.11\pm0.03$ & $-366.6\pm9.1$ \\
    CT-DQN ($\omega = 0.01$) & $75\pm17$ & $0.15\pm0.03$ & $-407.5\pm65.3$ \\
    CT-DQN ($\omega = 0.05$) & $70\pm3$ & $0.16\pm0.04$  & $-370.3\pm8.9$\\
    CT-DQN ($\omega = 0.1$) & $66\pm0.4$ & $0.15\pm0.04$ & {$-370.2\pm1.2$}\\
    \midrule
    \multicolumn{4}{c}{\textit{Lunar lander}}\\
    \noalign{\smallskip}
    DQN & N.A. & $0.18\pm0.11$ & $-82.5\pm10.22$ \\
    CT-DQN ($\omega=0.01$) & $\mathbf{431\pm55}$ & $0.09\pm0.01$ & $\mathbf{180.2\pm11.7}$\\
    CT-DQN ($\omega=0.05$) & $\mathbf{311\pm58}$ & $0.16\pm0.07$ & $\mathbf{189.8\pm24.1}$\\
    CT-DQN ($\omega=0.1$) & $\mathbf{529\pm82}$ & $0.14\pm0.09$ & $\mathbf{156.7\pm28.4}$\\
    \midrule
    \multicolumn{4}{c}{\textit{Car racing}}\\
    \noalign{\smallskip}
    DQN & - & - & $549.2 \pm 290$\\
    CT-DQN ($\omega = 0.05$) & - & - & $\mathbf{728 \pm 294.5}$\\
    \bottomrule
\end{tabular}
\caption{Control metrics (Def. \ref{def:control_metrics}) for the scenarios in § \ref{sec:evaluation}.
Means and standard deviations across sessions are reported, when $S > 1$.
Values that are statistically significantly different from those of DQN are in bold (according to Welch’s t-test with $p$-value less than $0.05$).}
\label{tab:control_metrics}
\end{table}

%-----------------------------------
\subsection{Lunar Lander} \label{sec:lunar_lander}

%-------------------------------
\paragraph{Environment description and control goal.}

In a 2-D space, a spaceship subject to gravity, in the absence of friction, must use its thrusters to land with reduced velocity on a landing pad.
The states are the coordinates and orientation of the lander, the corresponding velocities, and two Boolean variables to determine contact of the two legs with the ground.
The lander has three thrusters, on the left, on the right, and on the bottom (main) of the spacecraft. The possible (four) control inputs are the following: use only the left thruster, only the right one, the main one or no activation of any thruster. 
The position of the landing pad and the initial position and orientation of the lander are fixed, while the initial linear speed is random, as well as the terrain topography aside from the landing pad.
The spacecraft \emph{lands correctly} if it impacts on the pad with its legs at a moderate velocity, while it \emph{crashes} if its body touches the ground, or lands with a velocity that is too high.
Further detail can be found in \cite{GYM_lander}.
The agent obtains a high reward for landing correctly, a large negative one for crashing, and a small negative one for consuming fuel.
Following \cite{GYM_lander}, we set the goal condition as achieving $J^{\pi} \ge 200$ (i.e., $c_g$ true if $J^{\pi} \ge 200$).
It is worth noting that this might also be seen as a regulation problem, with the objective of reaching the center of the pad ($x^*$), in the origin of the reference frame.
Thus, we define the settling time $k_{\R{s}}$ as the instant when the spacecraft lands correctly, if it happens.
Moreover, we set $S=3$, $E=1000$, $N=1000$, although an episode ends immediately if the lander lands correctly or if it crashes.

%--------------------------------------------
\paragraph{Control tutor design.}

In order to design the tutor, we assume the knowledge of a simplified dynamics of the center of mass of the lander, by neglecting gravity. Indeed, its magnitude might be unknown.
Namely, we approximate $f$ in \eqref{eq:rl_problem_statement} with the reduced order model $\hat{f}\left(\chi_k, v_k \right) = A \chi_k + B v_k$,
where $\chi_k \in \BB{R}^4$ is the vector containing position and velocity on the x-axis followed by position and velocity on the y-axis (in this given order); $v_k \in \BB{R}^2$ are the x- and y- components of the force applied by a hypothetical swivelling thruster.
Noting that $\chi = 0$ corresponds to the center of the landing pad, we exploit the state-feedback control law defined as $v_k = -K \chi_k$ to stabilize asymptotically the origin, where $K \in \BB{R}^{2 \times 4}$.
The matrices of the reduced order model are defined as follows:

\begin{equation}
A = \left[ \begin{matrix} 
1 & T & 0 & 0\\
0 & 1 & 0 & 0\\
0 & 0 & 1 & T\\
0 & 0 & 0 & 1\\
\end{matrix} \right],
\quad 
B = \left[ \begin{matrix} 
0 & 0\\
T/m & 0\\
0 & 0\\
0 & T/m
\end{matrix} \right],
\quad
K = \left[ \begin{matrix} 
470 & 474.7 & 0 & 0\\
0 & 0 & 470 & 474.7
\end{matrix} \right],
\end{equation}
where $T = 0.02$ is a sampling time and $m = 10$ kg is the mass of the lander.
To obtain the control tutor's input $g(x_k)$ in \eqref{eq:tutor_policy} from $v_k(\chi_k)$, we proceed as follows.
If $v_y > 0$ and $|v_y| \ge |v_x|$ (the tutor mainly suggests moving upwards), we use the thruster on the bottom; if $|v_x| > |v_y|$ and $|v_x| > 0$ (the tutor mainly suggests moving right), we use the thruster on the left; if $|v_x| > |v_y|$ and $|v_x| < 0$ (the tutor mainly suggests moving left), we use the thruster on the right; in the other cases, we use no thruster.
Note that this control tutor, by itself, is unable to make the spacecraft land correctly as it has access only to a very limited amount of information on the system dynamics.

\begin{figure}[t]
\centering
\begin{minipage}{\textwidth}
\begin{tikzpicture}
  \node (img)  {\includegraphics[width=0.98\textwidth]{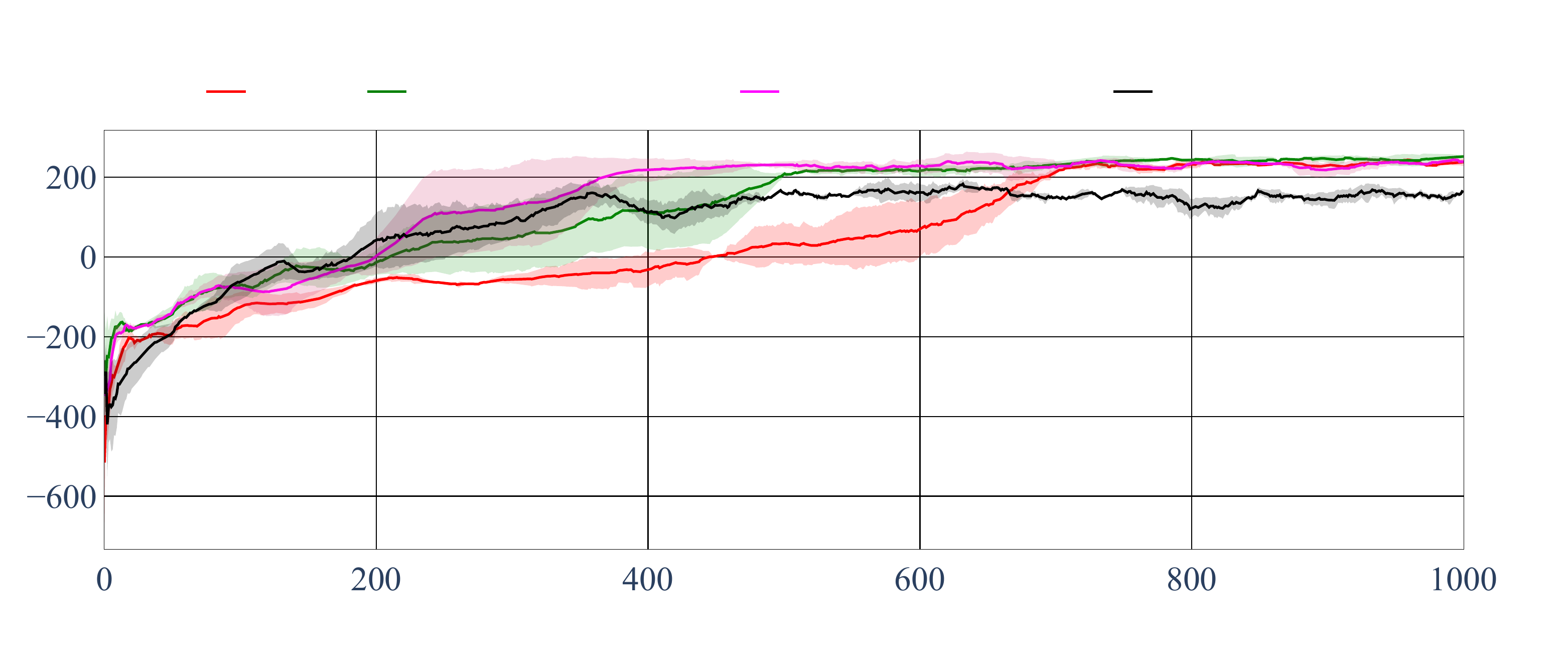}};
  \node[above=of img, node distance=0cm,xshift=-4.6cm,yshift=-2.3cm,font=\color{black}] {DQN};
  \node[above=of img, node distance=0cm,xshift=-2cm,yshift=-2.3cm,font=\color{black}] {CT-DQN $\omega =0.01$};
  \node[above=of img, node distance=0cm,xshift=1.5cm,yshift=-2.3cm,font=\color{black}] {CT-DQN $\omega =0.05$};
  \node[above=of img, node distance=0cm,xshift=5cm,yshift=-2.3cm,font=\color{black}] {CT-DQN $\omega =0.1$};
  \node[below=of img, node distance=0cm, yshift=1.6cm,font=\color{black}] {Episode number};
  \node[left=of img, node distance=0cm, rotate=90, anchor=center,yshift=-1cm,font=\color{black}] {Cumulative reward};
 \end{tikzpicture}
\end{minipage}
\caption{
Cumulative reward per episode $J^\pi_\R{e}$ for the lunar lander problem.
The reward curves were averaged with a moving window of 100 samples taken on the left.
Then mean (solid curves) and standard deviations (shaded areas) are calculated across sessions.
}
\label{fig:lander_rewards}
\end{figure}
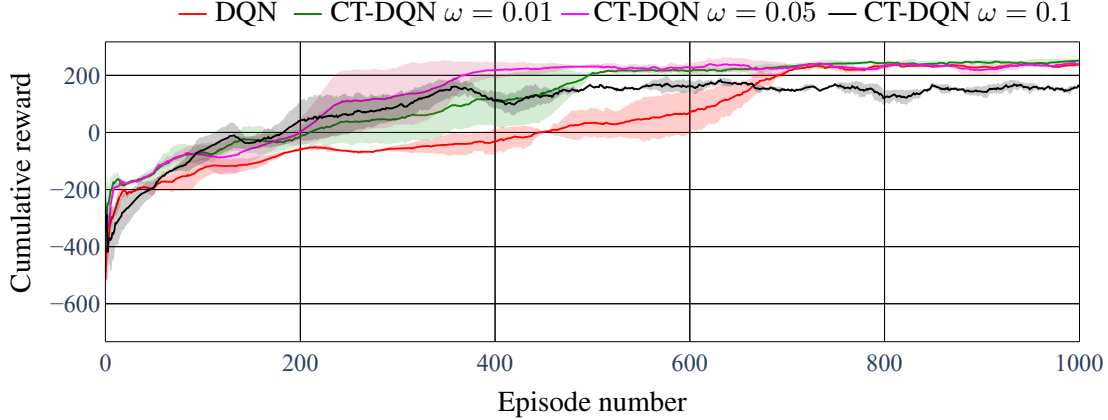

%------------------------------------------
\paragraph{Numerical results.}
Fig.~\ref{fig:lander_rewards} shows that CT-DQN improves the learning performance with respect to DQN, reducing learning times.
Notably, as reported in Tab.~\ref{tab:learning_metrics}, CT-DQN with $\omega = 0.05$ requires about half as many episodes as DQN to consistently achieve the goal (see $E_\R{t}$).
Also, the average cumulative reward across all episodes $J_{\R{avg}}^{\pi}$ of CT-DQN is more than twice that of DQN, indicating a shorter learning time.
Then, after both algorithms reach their their terminal episode $E_\R{t}$, they exhibit comparable average cumulative reward $J_{\R{avg,t}}^{\pi}$.
In Tab.~\ref{tab:control_metrics} we compare the control strategies $\pi_{\R{greedy}}$ obtained from CT-DQN and DQN by halting their training at $500$ episodes (after $500$ episodes, CT-DQN already converged, as its $E_{\R{t}} < 500$, while DQN has not, as its $E_{\R{t}} > 500$).
The DQN agent could not learn how to land yet, but keeps hovering over the landing pad, wasting fuel.
This is captured by the negative cumulative reward $J^{\pi_{\R{greedy}}}$, coupled with a low steady state error $e_{\R{s}}$, and the settling time $k_{\R{s}}$ being not available.
On the other hand, the CT-DQN agent has already learnt how to land, even with different values of $\omega$ (introduced in § \ref{sec:policy_design}), displaying a positive $J^{\pi_{\R{greedy}}}$, a finite $k_{\R{s}}$, and a low $e_{\R{s}}$.
We note also that, as the tutor is synthesized with only a partial model of the system dynamics, performance might start to degrade when the tutor is used \emph{too} often.
As evidence, see the asymptotic value of the reward curves of CT-DQN with $\omega = 0.1$ in Figs.~\ref{fig:pendulum_rewards} and~\ref{fig:lander_rewards}.

%---------------------------------------
\subsection{Car Racing}
\label{sec:car_racing}

%-------------------------------
\paragraph{Environment description and control goal.}

In a 2-D space, a car must complete a random track as fast as possible.
The state is composed of the pixel matrices of three consecutive image frames.
The actions are the possible combinations of ``steer left/right'', ``accelerate'', and ``brake'', all by a fixed amount.
The agent is rewarded positively each time it visits a new bit of road, and receives a small negative reward when time steps pass.
Further details are reported in \cite{GYM_car}.

In this case, rather than defining a specific goal which can or cannot be satisfied in an episode, we deem more natural to consider the task just as that of maximizing the reward; therefore, the only metric we consider are the cumulative rewards $J_{\R{avg}}^\pi$ and $J_{\R{avg}}^{\pi_\R{greedy}}$.
We set $S=2$, $E = 500$ and $N=1000$, although an episode can end earlier if the car gets too far from the track or visits $95\%$ of the track.

%-------------------------------------------
\paragraph{Control tutor design.}

\begin{figure}[t]
    \begin{minipage}{\textwidth}
    \centering
    \begin{tikzpicture}
        \node (img)  {\includegraphics{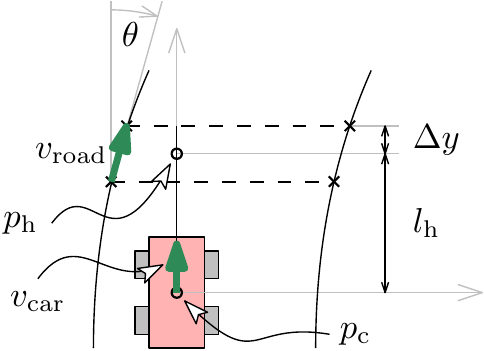}};
        \node[above=of img, node distance=0cm,xshift=-2.4cm,yshift=-1.8cm,font=\color{black}] {(a)};
        \qquad \qquad;
        \node[xshift=6cm] (img1) {\includegraphics{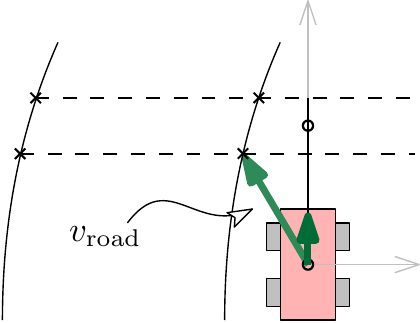}};
        \node[above=of img, node distance=0cm,xshift=3.6cm,yshift=-1.8cm,font=\color{black}] {(b)};
     \end{tikzpicture}
    \end{minipage}
    
    \caption{Quantities used by the control tutor, when the car is on the road (a) and off the road (b).}
    \label{fig:car_tutor}
\end{figure}

The tutor regulates acceleration and steering separately. 
Steering is regulated as follows.
First, we note that the car is still in each frame, is oriented upwards, and has its center of mass at position $p_{\R{c}} = [x_{\R{c}} \ \ y_{\R{c}}]\T = [0 \ \ 0]\T$ (in pixels) (Fig.~\ref{fig:car_tutor}).
Moreover, we detect the margins of the road by processing each image frame with a Roberts operator \citep{DAVIS1975248}.
Next, we consider a point in front of the car, with position $p_{\R{h}} = [x_{\R{h}} \ \ y_{\R{h}}]\T = [ x_{\R{c}} \ \ y_{\R{c}} + l_{\R{p}}]\T$, with $l_{\R{p}} \in \BB{R}_{> 0}$ (see Fig.~\ref{fig:car_tutor}).
We also consider two horizontal lines at $y_{\R{h}} + \Delta y$ and $y_{\R{h}} - \Delta y$, where $\Delta y \in \BB{R}{> 0}$.
Normally, these lines will intersect the margins of the road in four points (see again Fig.~\ref{fig:car_tutor}), and we define $v_{\R{road}}$ as the vector between the intersection points on the side of the road closer to $p_{\R{h}}$.
Let $\theta \coloneqq \angle(v_{\R{car}} - v_{\R{road}})$ be the angle of road with respect to the car.
Then, to align the car with the road, if $\theta < 0$ (resp. $\theta > 0$), the tutor suggests to steer left (resp.~right).
However, if all the intersection points are on one side with respect to $x_\R{c}$, or if less than four intersection points are found, it is inferred that the car is off the road, and $v_{\R{road}}$ is defined as the vector from $p_{\R{c}}$ to the closest intersection point, instead (see Fig. \ref{fig:car_tutor}.(b)).%
\footnote{More complicated situations might exist, e.g., where less than four intersection points are found, but the car is on the road; however, these are typically infrequent. We also do not aim to build the best possible tutor, but a simple one that is able to demonstrate the potential of the approach. Tutors that are able to provide the learning process with more accurate suggestions will lead to better performance. In a sense, a simple tutor can be considered a baseline over which improvements are possible.}
To regulate the speed $s$, first we detect $s$ by measuring an indicator bar printed on the image frame.
Then, setting some thresholds $\eta_{\R{speed}}^\R{acc}, \eta_{\R{angle}}^\R{acc}, \eta_{\R{angle}}^\R{brk} \in \BB{R}_{> 0}$,
the tutor suggests to accelerate if $s < \eta_{\R{speed}}^\R{acc}$ and $|\theta| < \eta_{\R{angle}}^\R{acc}$;
conversely, it suggests to brake if $|\theta| > \eta_{\R{angle}}^\R{brk}$.

%---------------------------------------------
\paragraph{Numerical results.} 

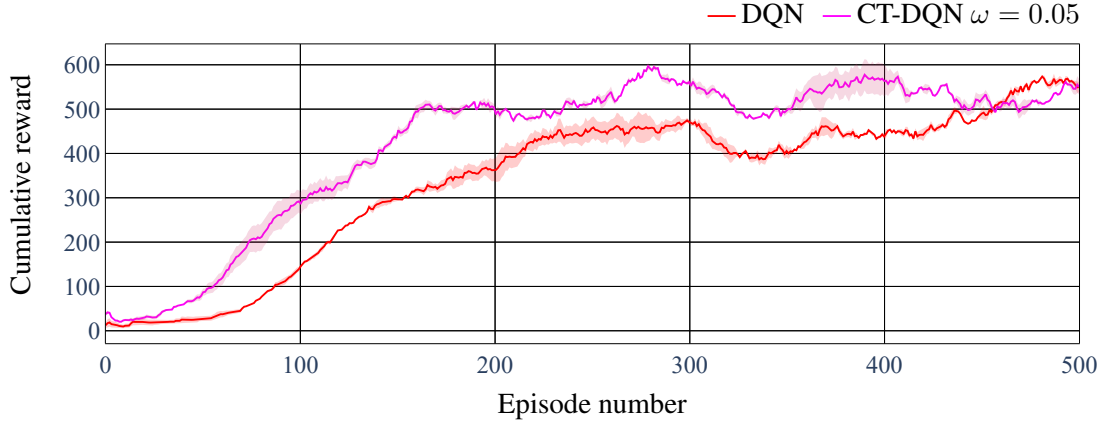
\begin{figure}[t]
\centering
\begin{minipage}{\textwidth}
\begin{tikzpicture}
  \node (img)  {\includegraphics[width=0.98\textwidth]{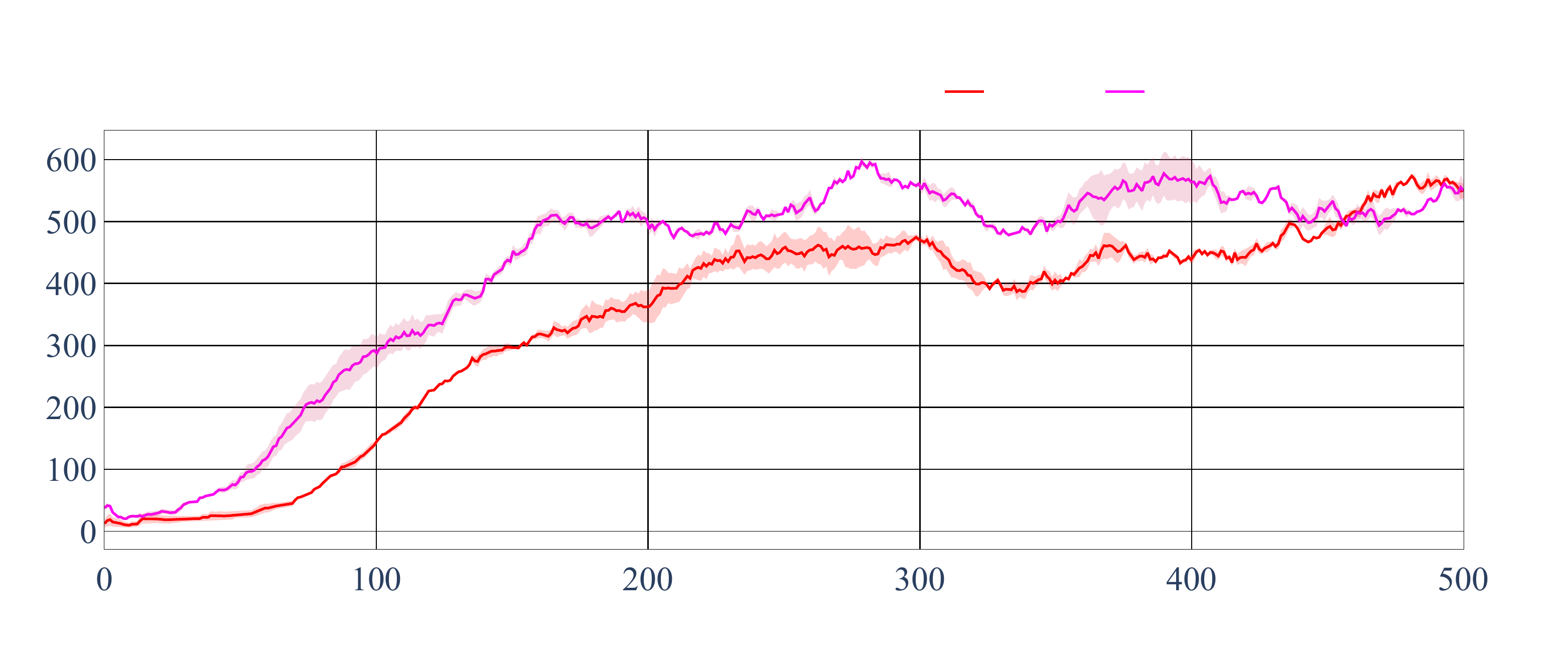}};
  \node[above=of img, node distance=0cm,xshift=2.4cm,yshift=-2.3
  cm,font=\color{black}] {DQN};
  \node[above=of img, node distance=0cm,xshift=5cm,yshift=-2.3
  cm,font=\color{black}] {CT-DQN $\omega=0.05$};
  \node[below=of img, node distance=0cm, yshift=1.6cm,font=\color{black}] {Episode number};
  \node[left=of img, node distance=0cm, rotate=90, anchor=center,yshift=-1cm,font=\color{black}] {Cumulative reward};
 \end{tikzpicture}
\end{minipage}
\caption{Cumulative reward per episode $J^\pi_e$ of DQN and CT-DQN.
The curves were obtained using a moving average of 50 samples (taken on the left).}
\label{fig:CR_rewards}
\end{figure}

Fig.~\ref{fig:CR_rewards} shows a generally faster learning for CT-DQN, as the cumulative reward is higher for almost the entire session.
This is also confirmed in Tab.~\ref{tab:learning_metrics} by the larger value of $J_{\R{avg}}^{\pi}$ for CT-DQN.
Additionally, we test the greedy policies obtained after training for $250$ episodes, on 30 tracks generated randomly (the same tracks for both algorithms).
We find significantly higher rewards (see $J^{\pi_{\R{greedy}}}$ in Tab.~\ref{tab:control_metrics}) for CT-DQN, showing the benefit of using the control tutor.

%----------------------------------
\section{Conclusions}

In this paper, we have presented Control-Tutored DQN (CT-DQN), a solution based on the integration of DRL algorithms with a tutor mechanism for aiding exploration based on control theory. In particular, we have discussed the design of an extended version of DQN, where actions are sometimes suggested by a controller (tutor).
In order to evaluate our approach, we have considered three representative scenarios of increasing complexity, i.e, the stabilization of an inverted pendulum, the landing of a spacecraft (i.e., lunar lander), and the control of a racing car.
In all the cases, we have considered tutors that rely on simple mechanisms and are designed with limited information about the systems dynamics.

We have shown that their addition always proved to be non-pejorative (with the inverted pendulum) or significantly beneficial (with the lunar lander and racing car) in terms of shorter learning time. 
We have also observed that we are able to obtain better policies with respect to classical DQN in the same number of episodes.
Moreover, the better the tutor is at solving a problem (according to case-specific metrics), the larger the improvement tends to be.
Our future agenda is focused around the formal analysis of the design of the tutor mechanism for Deep Reinforcement Learning, including the quantification of information and definition of bounds (e.g., regret bounds).

% REFERENCES %%%%%%%%%%%%%%%%%%%%%%%%%%%%%%%%%%%%%%%%%%%%%%%%%%%%%%%%%

\appendix

\section{Hyperparameters Tuning}
%\mm{This should go after the references in my opinion - at least this is typical in ML conferences.}
During training, we use a target neural network \citep{mnih2015human} which is updated at the end of every episode. 
We also introduce a replay buffer with size $N_{\R{b}}$, which is used to randomly sample $64$ data-points to update the network parameters at every step.
Moreover, we set learning rate $\alpha = 0.001$ and the discount factor $\gamma = 0.99$ \citep{Sutton1998}.
With respect to the inverted pendulum described in Section \ref{sec:inverted_pendulum},
%--------------------------------
% \subsection{Inverted Pendulum}
%
for the neural networks in DQN, we use $2$ hidden layers with rectifier linear unit activation functions (ReLu), with $128$ and $64$ nodes, respectively.
We set $N_{\R{b}} = 1,000,000$ and $\epsilon^{\R{rl}} = 0.02$.

As far as the lunar lander in Section \ref{sec:lunar_lander} is concerned,
%--------------------------------
% \subsection{Lunar Lander}
for the neural networks in DQN, we used $2$ hidden layers of $128$ nodes with ReLu.
Moreover, we set $N_{\R{b}} = 1,000,000$, $\alpha = 0.0001$, $\gamma = 0.99$, and $\epsilon{^\R{rl}} = 0.1$.
Finally for the car racing discussed in Section \ref{sec:car_racing},
%---------------------------------
% \subsection{Car Racing}
we use convolutional neural networks.
The input has dimension $94 \times 94 \times 3$.
A first hidden layer convolves $6$ filters of $7 \times 7$ with stride $3$ with the input image, with ReLu.
A second hidden layer convolves $12$ filters of $4 \times 4$ with stride $1$, with ReLu. 
A third hidden layer is present, with $216$ nodes and ReLu.
The output layer is a fully-connected linear layer with a
single output for each possible action.
Finally, we set $N_{\R{b}} = 5,000$, $\alpha = 0.001$, $\gamma = 0.9999$, and $\epsilon{^\R{rl}} = 0.1$, and select 
$l_{\R{p}} = 10$ pixels,
$\Delta y = 2$ pixels, $\eta_{\R{angle}}^\R{acc}=15^{\circ}$,  $\eta_{\R{angle}}^\R{brk}=50^{\circ}$, and $\eta_{\R{speed}}^\R{acc}$ as $40\%$ of the maximum possible speed. These values are selected as a representative scenario for this type of games.

% Acknowledgments---Will not appear in anonymized version
%\acks{We thank a bunch of people.}

\end{document}